\documentclass[11pt]{article}

\usepackage[margin=1in]{geometry}
\usepackage{amsmath, amssymb}
\usepackage{booktabs}
\usepackage{graphicx}
\usepackage{hyperref}
\usepackage{microtype}
\usepackage{caption}
\usepackage{placeins}   % FloatBarrier
\usepackage{float}      % H specifier if needed

\hypersetup{
  colorlinks=true,
  linkcolor=blue,
  citecolor=blue,
  urlcolor=blue
}

\graphicspath{{figures/}}

\title{Mitigating Position-Shift Failures in Text-Based Modular Arithmetic via Position Curriculum and Template Diversity}

\author{
  Nikolay Yudin \\
  Independent Researcher \\
  \texttt{n.yudin@gmail.com}
}

\date{06 January 2026}

\begin{document}
\maketitle

% ----------------------------
% Abstract (FINAL)
% ----------------------------
\begin{abstract}
Building on insights from the grokking literature \cite{power2022grokking,nanda2023progress,liu2022omnigrok}, we study character-level Transformers trained to compute modular addition from text, and focus on robustness under input-format variation rather than only in-distribution accuracy. We identify a previously under-emphasized failure mode: models that achieve high in-distribution accuracy can fail catastrophically when the same expression is shifted to different absolute character positions (“position shift”) or presented under out-of-distribution natural-language templates. Using a disjoint-pair split over all ordered pairs for p=97, we show that a baseline model reaches strong in-distribution performance yet collapses under position shift and template OOD. We then introduce a simple training recipe that combines (i) explicit expression boundary markers, (ii) position curriculum that broadens the range of absolute positions seen during training, (iii) diverse template mixtures, and (iv) consistency training across multiple variants per example. Across three seeds, this intervention substantially improves robustness to position shift and template OOD while maintaining high in-distribution accuracy, whereas an ALiBi-style ablation fails to learn the task under our setup. Our results suggest that steering procedural generalization under noisy supervision benefits from explicitly training invariances that are otherwise absent from the data distribution, and we provide a reproducible evaluation protocol and artifacts.
\end{abstract}

% ----------------------------
% 1. Introduction
% ----------------------------
\section{Introduction}

Modern neural networks can achieve near-perfect in-distribution performance while relying on brittle shortcuts that fail under small, realistic input-format shifts.
This gap is a core obstacle for reliable agents, instruction-following systems, and tool-using models, where supervision is inherently noisy (synthetic data, human labels, RLHF) and deployment inputs vary in phrasing, length, and structure.

Building on insights from the grokking literature, we study a controlled setting where a character-level Transformer is trained to compute modular addition from text.
Rather than optimizing only for in-distribution accuracy, we focus on robustness under input-format variation, and ask: \emph{can training be steered toward procedural solutions that are invariant to position and phrasing, rather than memorization tied to surface form?}

\paragraph{Connection to grokking.}
Classic grokking work studies a delayed transition from memorization to generalization under prolonged training and particular regularization conditions \cite{power2022grokking,nanda2023progress}.
We study a complementary axis: even when disjoint-pair generalization is strong, models may remain brittle to \emph{format} variation (absolute position shifts and out-of-distribution templates).
Our interventions can be viewed as \emph{steering invariances} that are absent from the training distribution, rather than waiting for spontaneous emergence of a robust procedure.

\paragraph{A minimal but revealing benchmark.}
We consider modular addition $(a+b)\bmod 97$ presented as text.
This task is algorithmic, yet admits shortcut learning: a model can perform well under a narrow training format while failing catastrophically when the same expression is shifted within the sequence or paraphrased.
We therefore evaluate with a small suite that separates in-distribution competence from procedural robustness:
(i) \textbf{Eval-A} (in-distribution), (ii) \textbf{Eval-B} (position shift), and (iii) \textbf{Eval-C0} (template OOD without anchors).
For anchor-based models we additionally report \textbf{Eval-C1} (template OOD with anchors).

\paragraph{Key observation: robustness collapses despite high in-distribution accuracy.}
A baseline model trained on a fixed, simple format achieves high in-distribution performance (Eval-A $96.8\pm 4.2$\%),
yet fails under position shift (Eval-B $14.9\pm 0.5$\%) and template OOD (Eval-C0 $1.2\pm 0.8$\%).
This ``robustness cliff'' indicates that the model does not learn a position- and format-invariant procedure, despite appearing successful on standard evaluation.

\paragraph{Steering interventions.}
We propose and ablate a set of training interventions aimed at forcing invariances:
(1) \textbf{position diversity} via padding-based control of the expression location,
(2) a \textbf{position curriculum} that gradually expands the allowed position range during training,
(3) \textbf{multi-variant training} with $K=4$ format variants per example and an explicit \textbf{consistency loss} to penalize disagreement across variants,
and (4) \textbf{template diversity} across padding-style and natural-language prompts.
We also study the use of lightweight \textbf{anchor tokens} (e.g., \texttt{<EXPR>...\ </EXPR>}) as optional structure markers, and evaluate both anchored and no-anchor test suites to avoid unfairly penalizing non-anchor baselines.

\paragraph{Main results.}
Position-only training already induces substantial robustness gains: I1\_001\_1 reaches Eval-B $71.7\pm 0.6$\% and Eval-C0 $60.3\pm 6.3$\% while maintaining Eval-A $96.5\pm 0.9$\%.
Our full intervention (I1\_002a) improves further, achieving Eval-B $73.7\pm 0.7$\% and Eval-C0 $80.5\pm 3.0$\% with Eval-A $96.0\pm 0.5$\%.
For anchor-based OOD evaluation, I1\_002a attains Eval-C1 $94.5\pm 2.2$\%.
In contrast, an ALiBi-based variant (I1\_002\_ALiBi) fails to learn the character-level parsing required in this setup (Eval-A $21.4\pm 1.0$\%).

\paragraph{Contributions.}
\begin{itemize}
  \item We demonstrate a \textbf{catastrophic position-shift failure mode} in character-level arithmetic-from-text: high in-distribution accuracy can coexist with near-random performance under moderate shifts.
  \item We introduce a \textbf{compact evaluation suite} (Eval-A/B/C0/C1) that separates in-distribution success from procedural robustness under format variation.
  \item We provide a \textbf{reproducible training recipe} (position curriculum + multi-variant consistency + template diversity, with optional anchors) that substantially improves robustness while preserving high in-distribution accuracy.
\end{itemize}

\paragraph{Paper structure.}
Section~\ref{sec:setup} defines the dataset, tokenization, model, and evaluation protocols.
Section~\ref{sec:methods} describes the interventions and training objectives.
Section~\ref{sec:results} reports results across three seeds and analyzes training dynamics.
Section~\ref{sec:discussion} discusses implications, limitations, and next steps.

% ----------------------------
% Related Work
% ----------------------------
\section{Related Work}
\label{sec:related_work}

\subsection{Grokking on algorithmic tasks and modular arithmetic}
Power et al.\ \cite{power2022grokking} introduced grokking as delayed generalization on small algorithmic datasets, highlighting the role of regularization.
Nanda et al.\ \cite{nanda2023progress} proposed progress measures and mechanistic perspectives on grokking dynamics.
Liu et al.\ \cite{liu2022omnigrok} studied grokking beyond strictly algorithmic data, suggesting that diversity of formats can influence generalization.
In this context, our focus is not the timing of delayed generalization per se, but a robustness failure mode that can persist despite strong disjoint-pair performance.

\subsection{Steering or accelerating grokking-like transitions}
Several works propose methods to accelerate grokking or characterize its geometry, e.g., gradient-based acceleration \cite{lee2024grokfast} and Jacobian/alignment-based regularization \cite{walker2025grokalign}.
We do not aim to accelerate grokking in the classic sense; instead, we explicitly penalize format-specific solutions via multi-variant consistency and curated format diversity.
These approaches appear complementary and could be combined in future work.

\subsection{Grokking under realistic conditions and failure modes}
Recent studies investigate grokking in more realistic settings and broader domains \cite{liu2025grokkingwild,chen2025grokkingpretraining}.
Relatedly, work on generalization collapse emphasizes that improvements on one axis can coincide with failures on another \cite{prakash2025generalizationcollapse}.
Our ``position shift'' and template-OOD failures instantiate a concrete, testable version of this concern for text-based algorithmic supervision.

\subsection{Distribution shift and domain adaptation in NLP}
Robustness to distribution shift has a long history in NLP and ML, including classic domain adaptation settings and theory-motivated bounds.
Modern neural NLP systems often rely on data augmentation, invariance objectives, and domain-adaptive pretraining to improve transfer across domains and styles.
Our evaluation suite instantiates a structured, controllable shift: the underlying function is fixed, but surface form changes via absolute position and prompt templates.
The I1 recipe can be viewed as explicit invariance training (position/template augmentation + consistency regularization), aligned with the broader goal of improving reliability under format and domain shift.

% ----------------------------
% 2. Setup
% ----------------------------
\section{Setup}
\label{sec:setup}

\subsection{Task and dataset}
\label{sec:task}

We study modular addition over a prime modulus $p=97$.
Each example is defined by an ordered pair $(a,b)$ where $a,b \in \{0,\dots,p-1\}$, with label
\[
y = (a+b)\bmod p \in \{0,\dots,p-1\}.
\]
The full universe contains $p^2 = 9409$ ordered pairs.

\paragraph{Disjoint-pair split.}
To prevent pair-level memorization, we use a \emph{disjoint-pair} split: all 9409 ordered pairs are shuffled with a fixed seed and split 50/50 into train and test.
This yields 4704 training pairs and 4705 test pairs (no overlap).
Unless stated otherwise, all reported results aggregate over three random seeds $\{42,43,44\}$, where the seed controls the pair shuffle and training randomness.

\subsection{Text rendering and position definition}
\label{sec:text_rendering}

Inputs are rendered as character sequences containing (i) optional prefix padding and/or natural-language text, (ii) an arithmetic expression for the same underlying pair $(a,b)$, and (iii) optional expression boundary markers (anchors).

\paragraph{Position.}
We define the \emph{expression position} as the absolute character index of the \emph{first digit} of the first number in the rendered expression (i.e., the position of the first digit of $a$, not the position of an anchor token).
Padding is implemented by prepending filler text to move the expression to a target position.

\subsection{Tokenizer and vocabulary}
\label{sec:tokenizer}

We use a character-level tokenizer with a fixed vocabulary of size 80 and maximum sequence length 100.
The vocabulary includes digits, basic punctuation and whitespace, and a subset of Latin letters sufficient to express the template set.
We also include special tokens for padding and classification; for anchor-based experiments we additionally include explicit boundary markers (e.g., \texttt{<EXPR>} and \texttt{</EXPR>}) as atomic tokens.
All sequences are padded/truncated to the maximum length.

\subsection{Model}
\label{sec:model}

All experiments use the same small Transformer classifier:
a 2-layer Transformer encoder with $d_{\text{model}}=128$, $n_{\text{heads}}=4$, learned absolute positional embeddings up to length 100, and CLS pooling.
A learned CLS token is prepended to the embedded sequence; the final prediction is produced by a linear classifier from the CLS representation to $p$ classes.
Unless stated otherwise, we use learned absolute positional embeddings; one ablation replaces them with an ALiBi-style relative bias.

\subsection{Training}
\label{sec:training}

All runs are trained for a fixed budget of 5000 optimizer steps with batch size 256, AdamW optimizer, learning rate $10^{-3}$, and weight decay 0.01.
We report mean and standard deviation over three seeds.

\paragraph{Multi-variant training and consistency loss.}
For experiments with $K>1$, each underlying pair $(a,b)$ is rendered into $K$ distinct textual variants (e.g., different positions and/or templates).
We minimize a joint objective
\[
\mathcal{L} = \mathcal{L}_{\text{CE}} + \lambda \,\mathcal{L}_{\text{cons}},
\]
where $\mathcal{L}_{\text{CE}}$ is standard cross-entropy on the correct class for each variant, and $\mathcal{L}_{\text{cons}}$ encourages agreement across variants of the same pair.
Concretely, we compute pairwise mean-squared error between the pre-softmax logits across the $K$ variants and average over all pairs; we use $\lambda=1.0$ when consistency loss is enabled.

\paragraph{Position curriculum (steps-based).}
For position-diverse experiments, we apply a steps-based curriculum that gradually expands the allowed position range:
\begin{itemize}
  \item steps 0--1666: target position range $[10,30]$
  \item steps 1667--3333: target position range $[10,50]$
  \item steps 3334--5000: target position range $[10,70]$
\end{itemize}

\paragraph{Template diversity.}
When enabled, templates are sampled from a mixture of (i) padding-style templates, (ii) natural-language templates, and (iii) mixed templates (padding plus natural language).
We use a 40/40/20 mixture (padding / natural-language / mixed).

\subsection{Evaluation protocols}
\label{sec:eval_protocols}

We evaluate robustness with a compact suite that separates in-distribution competence from invariance to format shifts.

\paragraph{Eval-A (in-distribution generalization).}
Eval-A measures accuracy on a fixed set of $n=400$ pairs sampled from the test split (the same 400 pairs for all experiments).
Inputs are rendered using the evaluation template set corresponding to the experiment family, without adversarial position shifting.

\paragraph{Eval-B (position shift).}
Eval-B measures accuracy when the same arithmetic expression is shifted to different absolute character positions.
We evaluate at fixed target positions
\[
\{0, 8, 16, 24, 32, 48, 64\},
\]
and generate $n=100$ random pairs per position, yielding 700 total examples.

\paragraph{Eval-C0 (template OOD, no-anchor).}
Eval-C0 measures robustness to out-of-distribution natural-language templates \emph{without anchors}.
We generate $n=200$ examples using OOD templates split across two categories (\emph{questions} and \emph{commands}); positions are randomized via padding to cover the evaluation range.
Eval-C0 is applied to \emph{all} experiments.

\paragraph{Eval-C1 (template OOD, anchor).}
Eval-C1 mirrors Eval-C0 but includes explicit anchors around the expression (e.g., \texttt{<EXPR>...\ </EXPR>}).
We report Eval-C1 only for experiments trained with anchors; for other experiments Eval-C1 is not defined.

\paragraph{ConsistencyCorrect@4.}
For experiments trained with multi-variant inputs ($K=4$), we report \texttt{ConsistencyCorrect@4}: the fraction of evaluated pairs for which all four variants yield the same prediction and that prediction is correct.

% ----------------------------
% 3. Steering Interventions (I1)
% ----------------------------
\section{Steering Interventions (I1)}
\label{sec:methods}

Our goal is to steer models away from brittle, surface-form shortcuts and toward solutions that are invariant to input position and phrasing.
We implement this as a small set of training-time interventions that explicitly enforce invariances which are absent from a narrow training distribution.

\subsection{Intervention components}

\paragraph{(C1) Position diversity via controlled padding.}
We explicitly control the absolute character position of the arithmetic expression by prepending filler text.
For each example we sample a target position $t$ in a specified range and generate a prefix whose length places the first digit of the first operand at position $t$.

\paragraph{(C2) Steps-based position curriculum.}
We train with a curriculum that gradually expands the allowed position range over training steps.

\paragraph{(C3) Template diversity.}
We sample from a mixture of template families: padding-style, natural-language, and mixed (40/40/20).

\paragraph{(C4) Expression boundary markers (anchors).}
When enabled, we wrap the arithmetic expression with explicit boundary markers (\texttt{<EXPR>}, \texttt{</EXPR>}) inserted as atomic tokens in the character vocabulary.

\paragraph{(C5) Multi-variant training with consistency loss.}
For settings with $K>1$, each underlying pair $(a,b)$ is rendered into $K$ variants that differ in position and/or template while sharing the same label.
We add a consistency loss that penalizes disagreement across variants of the same pair.

\subsection{Experiment matrix}

\paragraph{Baseline-001 (no steering).}
No padding, no template diversity, no anchors, $K=1$.

\paragraph{I1-001.1 (position steering).}
Position diversity + curriculum, $K=4$ + consistency loss, no template diversity, no anchors.

\paragraph{I1-002a (full steering).}
Position diversity + curriculum, template diversity, anchors enabled, $K=4$ + consistency loss.

\paragraph{I1-002-ALiBi (ablation).}
Same as I1-002a but using an ALiBi-style attention bias in place of learned absolute positional embeddings.

% ----------------------------
% 4. Results
% ----------------------------
\section{Results}
\label{sec:results}

We report mean $\pm$ standard deviation over three seeds $\{42,43,44\}$.
All runs are trained for 5000 optimizer steps.
Our goal is to separate (i) disjoint-pair generalization in-distribution (Eval-A) from
(ii) robustness to absolute position shifts (Eval-B) and (iii) robustness to OOD templates (Eval-C0/C1).

\subsection{Overall performance on the evaluation suite}
\label{sec:results_overall}

Table~\ref{tab:main_metrics} summarizes the main metrics and highlights a sharp robustness gap.

\begin{table}[t]
\centering
\caption{Main evaluation metrics (mean $\pm$ std over three seeds). Eval-A: in-distribution disjoint-pair generalization. Eval-B: position-shift robustness. Eval-C0: template OOD without anchors. Eval-C1: template OOD with anchors (reported only for anchor-trained models). ConsistencyCorrect@4 is reported only for $K{=}4$ multi-variant training.}
\label{tab:main_metrics}
\begin{tabular}{lccccc}
\toprule
Model & Eval-A & Eval-B & Eval-C0 & Eval-C1 & ConsistencyCorrect@4 \\
\midrule
Baseline-001     & 96.8$\pm$4.2 & 14.9$\pm$0.5 & 1.2$\pm$0.8  & \textemdash & \textemdash \\
I1\_001\_1        & 96.5$\pm$0.9 & 71.7$\pm$0.6 & 60.3$\pm$6.3 & \textemdash & 94.2$\pm$1.7 \\
I1\_002a          & 96.0$\pm$0.5 & 73.7$\pm$0.7 & 80.5$\pm$3.0 & 94.5$\pm$2.2 & 93.9$\pm$0.8 \\
I1\_002-ALiBi     & 21.4$\pm$1.0 & 34.3$\pm$3.0 & 15.5$\pm$2.2 & 34.5$\pm$3.5 & 20.3$\pm$1.1 \\
\bottomrule
\end{tabular}
\end{table}

The baseline attains high in-distribution accuracy (Eval-A $96.8\pm 4.2$\%) but collapses under both position shift (Eval-B $14.9\pm 0.5$\%) and no-anchor template OOD (Eval-C0 $1.2\pm 0.8$\%).
Position-only steering (I1\_001\_1) closes most of this gap (Eval-B $71.7\pm 0.6$\%, Eval-C0 $60.3\pm 6.3$\%) while maintaining Eval-A $96.5\pm 0.9$\%.
The full intervention (I1\_002a) improves template robustness further (Eval-C0 $80.5\pm 3.0$\%) while preserving high Eval-A ($96.0\pm 0.5$\%) and strong position-shift robustness (Eval-B $73.7\pm 0.7$\%); on anchor OOD, it reaches Eval-C1 $94.5\pm 2.2$\%.
In contrast, the ALiBi ablation fails to learn the task (Eval-A $21.4\pm 1.0$\%) and remains low across protocols.

\begin{figure}[t]
  \centering
  \includegraphics[width=\linewidth]{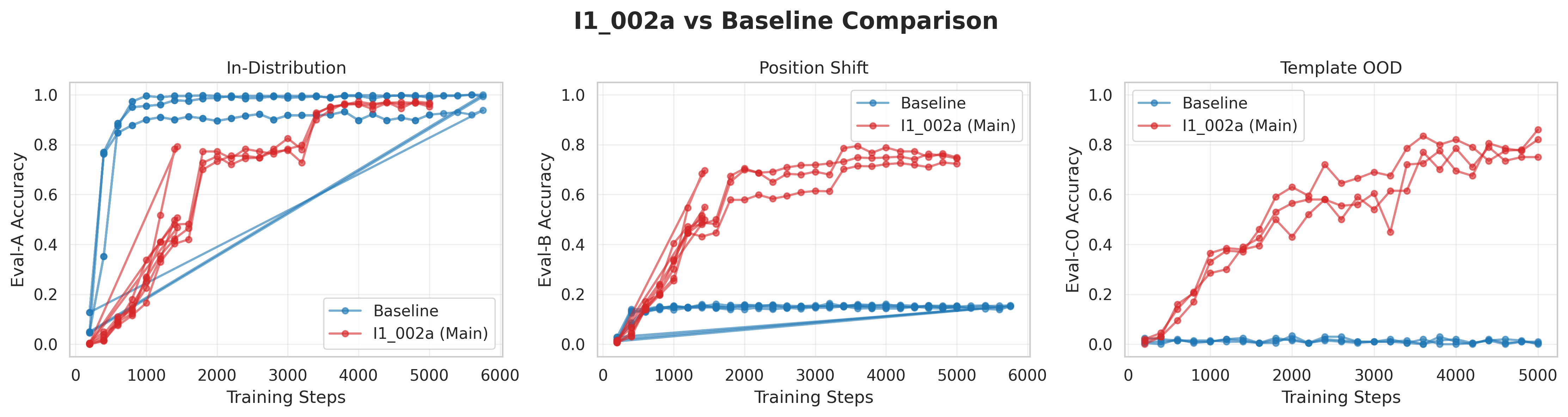}
  \caption{Baseline vs.\ I1\_002a on the evaluation suite (aggregated over seeds). The key gap is robustness: Eval-B and Eval-C0 improve substantially under I1.}
  \label{fig:comparison_i1_002a_vs_baseline}
\end{figure}

Figure~\ref{fig:comparison_i1_002a_vs_baseline} visualizes the same story: robustness gains are not marginal,
but a qualitative shift from near-random OOD behavior to substantially stable performance under format variation.

\FloatBarrier

\subsection{Training dynamics and when robustness emerges}
\label{sec:results_dynamics}

\begin{figure}[t]
  \centering
  \includegraphics[width=\linewidth]{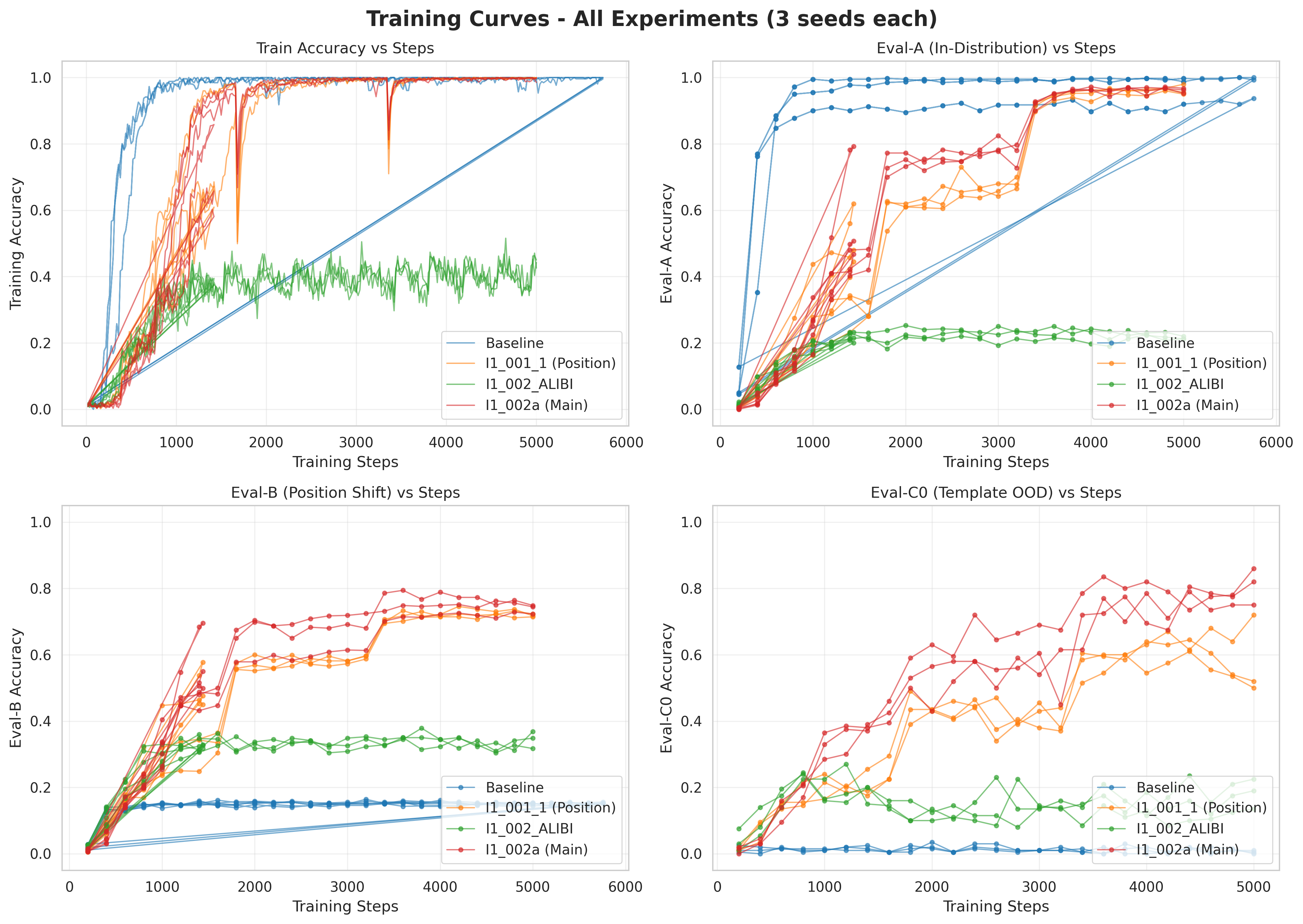}
  \caption{Training curves across experiments (aggregated view). Curves illustrate that in-distribution success can coexist with robustness collapse, and that invariance training changes the outcome.}
  \label{fig:training_curves_all}
\end{figure}

Figure~\ref{fig:training_curves_all} shows training curves across all experiments.
A key takeaway is that in-distribution success can arise without invariance: the baseline reaches high Eval-A yet retains poor robustness (Eval-B and Eval-C0).
By contrast, I1 training introduces multi-variant supervision and a position curriculum, and the improvement is reflected not only in final scores but also in higher agreement across variants (I1\_002a achieves ConsistencyCorrect@4 $93.9\pm 0.8$\%; Table~\ref{tab:main_metrics}), consistent with a shift toward format-invariant solutions.

\FloatBarrier

\subsection{Position-shift failure mode and mitigation}
\label{sec:results_position_shift}

\begin{figure}[t]
  \centering
  \includegraphics[width=\linewidth]{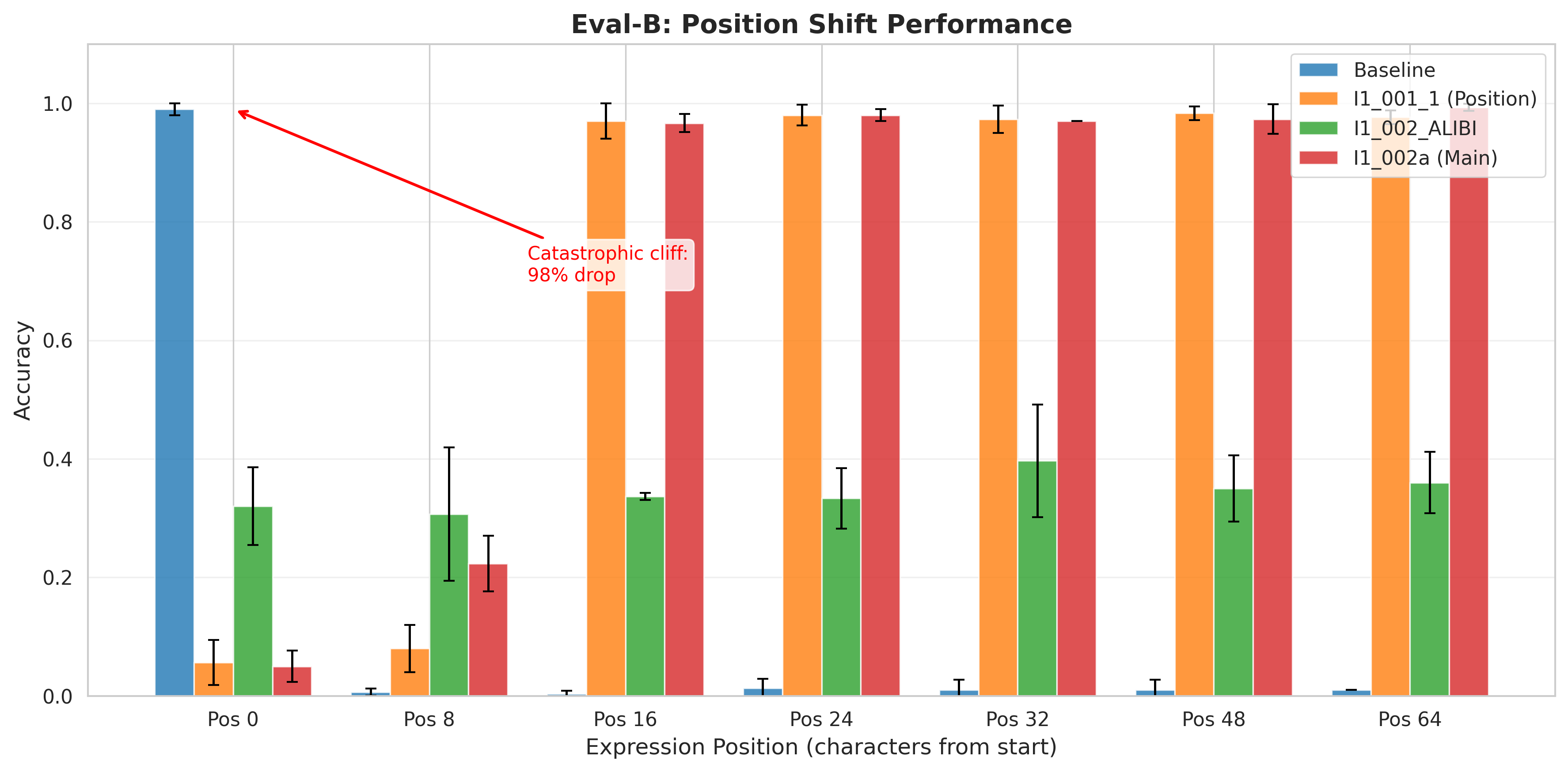}
  \caption{Eval-B position breakdown (aggregated over seeds). The baseline collapses under moderate shifts; position curriculum removes the cliff over the trained position range.}
  \label{fig:position_breakdown}
\end{figure}

The defining failure mode is a sharp drop in accuracy when the same expression is shifted to different absolute character positions.
Table~\ref{tab:pos_breakdown} 
\begin{table}[t]
\centering
\caption{Eval-B accuracy (\%) by absolute expression start position (mean $\pm$ std over three seeds). Positions 0 and 8 are intentionally outside the position curriculum (which starts at 10) and serve as OOD stress tests.}
\label{tab:pos_breakdown}
\small
\begin{tabular}{lccccccc}
\toprule
Model & pos0 & pos8 & pos16 & pos24 & pos32 & pos48 & pos64 \\
\midrule
Baseline-001   & 99.0$\pm$1.0 & 0.7$\pm$0.6 & 0.3$\pm$0.6 & 1.3$\pm$1.5 & 1.0$\pm$1.7 & 1.0$\pm$1.7 & 1.0$\pm$0.0 \\
I1\_001\_1      & 5.7$\pm$3.8  & 8.0$\pm$4.0 & 97.0$\pm$3.0 & 98.0$\pm$1.7 & 97.3$\pm$2.3 & 98.3$\pm$1.2 & 97.7$\pm$1.2 \\
I1\_002a        & 5.0$\pm$2.6  & 22.3$\pm$4.7 & 96.7$\pm$1.5 & 98.0$\pm$1.0 & 97.0$\pm$0.0 & 97.3$\pm$2.5 & 99.3$\pm$0.6 \\
I1\_002-ALiBi   & 32.0$\pm$6.6  & 30.7$\pm$11.2 & 33.7$\pm$0.6 & 33.3$\pm$5.1 & 39.7$\pm$9.5 & 35.0$\pm$5.6 & 36.0$\pm$5.2 \\
\bottomrule
\end{tabular}
\end{table}

and Figure~\ref{fig:position_breakdown} break down Eval-B by target position.
The baseline exhibits a catastrophic cliff: it is strong at pos~0 ($99.0\pm 1.0$\%) but drops to near-random at pos~8 ($0.7\pm 0.6$\%) and remains near chance for larger shifts (e.g., pos~16: $0.3\pm 0.6$\%, pos~64: $1.0\pm 0.0$\%).
In contrast, position-diverse training removes this cliff in the trained range: I1\_001\_1 sustains high accuracy at pos~16 ($97.0\pm 3.0$\%) and beyond (pos~64: $97.7\pm 1.2$\%).
The full intervention I1\_002a retains this benefit (pos~16: $96.7\pm 1.5$\%, pos~64: $99.3\pm 0.6$\%) while improving template OOD robustness (Section~\ref{sec:results_templates}).

Notably, positions 0 and 8 are intentionally outside the curriculum (training starts at position 10), so lower scores there
(e.g., I1\_002a pos~0: $5.0\pm 2.6$\%; pos~8: $22.3\pm 4.7$\%) should be interpreted as expected OOD stress tests rather than a failure of the intervention.

\FloatBarrier

\subsection{Template OOD robustness and the role of anchors}
\label{sec:results_templates}

Beyond position shift, we test robustness to out-of-distribution natural-language templates.
I1\_002a improves substantially on no-anchor OOD (Eval-C0 $80.5\pm 3.0$\% vs.\ baseline $1.2\pm 0.8$\%), indicating that robustness is not limited to anchored prompts.
For anchor-trained models, anchor OOD is also strong (I1\_002a Eval-C1 $94.5\pm 2.2$\%), suggesting that explicit boundary markers can act as a lightweight structural prior while still transferring to no-anchor evaluation.

\begin{figure}[t]
  \centering
  \includegraphics[width=\linewidth]{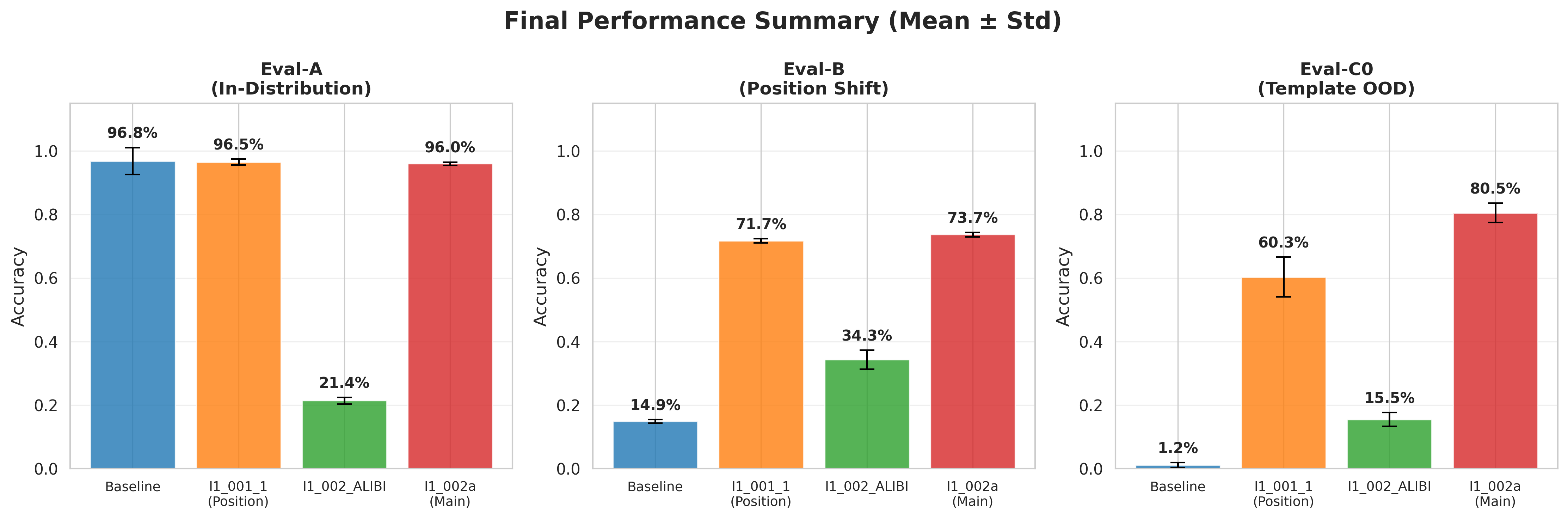}
  \caption{Final performance summary across protocols (aggregated over seeds). I1\_002a improves no-anchor OOD (Eval-C0) while also performing strongly on anchor OOD (Eval-C1).}
  \label{fig:final_performance_summary}
\end{figure}
\FloatBarrier

% ----------------------------
% 5. Discussion and Limitations (paper-ready)
% ----------------------------
\section{Discussion and limitations}
\label{sec:discussion}

\paragraph{Robustness is not implied by disjoint-pair generalization.}
The central lesson is that strong disjoint-pair generalization can coexist with catastrophic format brittleness.
Baseline-001 achieves high in-distribution accuracy (Eval-A $96.8\pm 4.2$\%) yet collapses under position shift (Eval-B $14.9\pm 0.5$\%) and no-anchor template OOD (Eval-C0 $1.2\pm 0.8$\%; Table~\ref{tab:main_metrics}), consistent with reliance on surface-form shortcuts rather than format-invariant procedures (Section~\ref{sec:results_overall}).

\paragraph{Position shift exposes a concrete robustness cliff.}
Eval-B reveals an extreme failure mode: the baseline is strong at pos~0 ($99.0\pm 1.0$\%) but drops to near-random at pos~16 ($0.3\pm 0.6$\%) and remains near chance at larger shifts (e.g., pos~64: $1.0\pm 0.0$\%; Table~\ref{tab:pos_breakdown}, Section~\ref{sec:results_position_shift}).
Position-diverse training with a curriculum removes this cliff over the trained range (e.g., I1\_001\_1 pos~16 $97.0\pm 3.0$\%; I1\_002a pos~16 $96.7\pm 1.5$\%; Table~\ref{tab:pos_breakdown}), suggesting that robustness requires explicitly expanding support over nuisance factors that are absent from the baseline distribution.

\paragraph{Template OOD and anchors as a lightweight structural prior.}
Template robustness improves substantially only when training includes explicit template diversity: I1\_002a reaches Eval-C0 $80.5\pm 3.0$\% versus $60.3\pm 6.3$\% for position-only steering (I1\_001\_1), and far above the baseline (Table~\ref{tab:main_metrics}, Section~\ref{sec:results_templates}).
Anchors provide an additional structural prior: for anchor-trained models, Eval-C1 is high (I1\_002a $94.5\pm 2.2$\%) while anchor-free OOD remains strong (Eval-C0), indicating that anchors do not merely ``move the goalposts'' but can coexist with no-anchor robustness.

\paragraph{Negative result: removing learned positional embeddings can prevent learning.}
The ALiBi-style ablation~\cite{press2022alibi} fails to learn the task under our character-level parsing setup (Eval-A $21.4\pm 1.0$\%; Table~\ref{tab:main_metrics}), suggesting that learned absolute positional embeddings can be functionally useful for span localization and segmentation in this regime (multi-digit grouping and delimiter structure), and that robustness here is achieved by training invariances rather than removing positional signals.

\paragraph{Limitations and next steps.}
The task is intentionally narrow (single-step modular addition), and the robustness suite covers specific template families and position ranges; in particular, the position curriculum excludes positions 0--8, which remain deliberate OOD stress tests (Section~\ref{sec:results_position_shift}).
Future work should extend to compositional expressions, broader natural-language variation, and mechanistic analyses of the learned invariances.
Methodologically, it is natural to combine explicit invariance training with grokking-acceleration techniques \cite{lee2024grokfast,walker2025grokalign} to improve both convergence speed and robustness under noisy supervision.

\paragraph{Outlook.}
Overall, these results provide a compact, reproducible testbed for \emph{steering procedural generalization under noisy supervision} by directly training invariances to realistic format variation.

% ----------------------------
% 7. Conclusion
% ----------------------------
\section{Conclusion}
\label{sec:conclusion}

We studied character-level Transformers trained to compute modular addition from text under a disjoint-pair split, focusing on robustness to input-format variation.
We identified a sharp position-shift failure mode: high in-distribution accuracy can coexist with near-random performance under modest shifts and OOD prompt templates.
A simple steering recipe---position curriculum, multi-variant consistency training, and template diversity (with optional anchors)---substantially improves robustness to both position shift and template OOD while preserving high in-distribution performance.
An ALiBi-style ablation fails to learn the task under our setup, suggesting that robustness here is achieved by explicitly training invariances rather than removing positional signals.
We release a reproducible evaluation protocol and artifacts to support further work on steering procedural generalization under noisy supervision.

% ----------------------------
% Reproducibility statement (optional)
% ----------------------------
\section*{Reproducibility}
We provide:
\begin{itemize}
\item Complete experiment configurations (architecture, optimizer, curriculum schedules)
\item Training scripts (\texttt{unified\_paper\_experiment\_v2.py})
\item Evaluation protocols (Eval-A/B/C0/C1 with sampling details)
\item Aggregated results (\texttt{final\_paper\_results.json}, 
      \texttt{reproducibility\_package.json})
\item Training curves for all 12 runs (4 experiments × 3 seeds)
\item Plotting code (\texttt{plot\_paper\_figures.ipynb} for Colab)
\item Documentation (\texttt{README\_REPRODUCIBILITY.md}, 
      \texttt{EXPERIMENT\_SUMMARY.md})
\end{itemize}
All artifacts are available at: 
\url{https://github.com/nick-yudin/Generalization/tree/main/papers/Mitigating_Position-Shift_Failures}.

\end{document}